# Connecting Distant Entities with Induction through Conditional Random Fields for Named Entity Recognition: Precursor-Induced CRF

**Wangjin Lee**[1] and **Jinwook Choi**[1,2,3]*

[1] Interdisciplinary Program for Bioengineering, Seoul National University, South Korea
[2] Department of Biomedical Engineering, College of Medicine, Seoul National University, South Korea
[3] Institute of Medical and Biological Engineering, Medical Research Center, Seoul National University, South Korea
`{jinsamdol,jinchoi}@snu.ac.kr`

**Abstract**

This paper presents a method of designing specific high-order dependency factor on the linear chain conditional random fields (CRFs) for named entity recognition (NER). Named entities tend to be separated from each other by multiple outside tokens in a text, and thus the first-order CRF as well as the second-order CRF may innately lose transition information between distant named entities. The proposed design uses outside label in NER as a transmission medium of precedent entity information on the CRF. Then, empirical results apparently demonstrate that it is possible to exploit long-distance label dependency in the original first-order linear chain CRF structure upon NER while reducing computational loss rather than in the second-order CRF.

## 1 Introduction

The concept of conditional random fields (CRFs) (John Lafferty, Andrew McCallum, & Fernando Pereira, 2001) has been successfully adapted in many sequence labeling problems (Andrew McCallum & Wei Li, 2003; Fei Sha & Fernando Pereira, 2003; John Lafferty et al., 2001; McDonald & Pereira, 2005). Even in deep-learning architecture, CRF has been used as a fundamental element in named entity recognition (Lample, Ballesteros, Subramanian, Kawakami, & Dyer, 2016; Liu, Tang, Wang, & Chen, 2017).

One of the primary advantages of applying the CRF to language processing is that it learns transition factors between hidden variables corresponding to the label of single word. The fundamental assumption of the model is that the current hidden state is conditioned on present observation as well as the previous state. For example, a part-of-speech (POS) tag depends on the word itself, as well as the POS tag transitions from the previous word. In the problem, the POS tags are adjacent to each other in a text forming a tag sequence; therefore, the sequence labeling model can fully capture dependencies between labels.

In contrast, a CRF in named entity recognition (NER) cannot fully capture dependencies between named entity (NE) labels. According to Ratinov & Roth (2009), named entities in a text are separated by successive "outside tokens" (i.e., words that are non-named entities syntactically linking two NEs) and considerable number of NEs have a tendency to exist at a distance from each other. Therefore, high-order interdependencies of named entities between successive *outside* tokens are not captured by first-order or second-order transition factors.

One major issue in previous studies was concerned with the way in which to explore long-distance dependencies in NER. Only dependencies between neighbor labels are generally used in practice because conventional high-order CRFs are known to be intractable in NER (Ye, Lee, Chieu, & Wu, 2009). Previous studies have demonstrated that implementation of the higher-order CRF exploiting pre-defined label patterns leads to slight performance improvement in the conventional CRF in NER (Cuong, Ye, Lee, & Chieu, 2014; Fersini, Messina, Felici, & Roth, 2014; Sarawagi & Cohen, 2005; Ye et al., 2009). However, there are certain drawbacks associated with handling named entity transitions within arbitrary length outside tokens.

In an attempt to utilize long-distance transition information of NEs through non-named entity to-kens, this study explores the method which modifies the first-order linear-chain CRF by using the induction method.

## 2 Precursor-induced CRF

Prior to introducing the new model formulation, the following information presents the general concept of CRF. As a sequence labeling model, the conventional CRF models the conditional distribution $P(y|x)$ in which $x$ is the input (e.g., token, word) sequence and $y$ is the label sequence of $x$. A hidden state value set consists of target entity labels and a single *outside* label. By way of illustration, presume a set $\{A, B, O\}$ as the hidden state value set; assign $A$ or $B$ to NEs, likewise, assign $O$ to outside words. From the hidden state set, a label sequence is formed in a linear chain in NER; for example, a sequence $\langle A, O, \cdots O, B \rangle$ in which successive outside words are between the two NE words. Because the first-order model assumes that state transition dependencies exist only between proximate two labels to prevent an increase in computational complexity, the first-order CRF learns bigram label transitions from the subsequence; $\{(A,O), (O,O), (O,B)\}$ that is, label transition data learnt from the example sequence. In the example, dependency $(A, B)$ is not captured in the model.

The main purpose of the precursor-induced CRF model, introduced in this study, is to capture specific high-order named entity dependency that is an outside word sequence between two NEs. The main idea can be explained in the following manner:

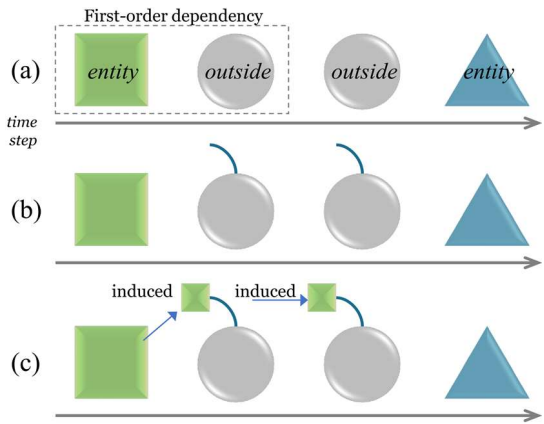

Figure 1: Transformation from conventional CRF to precursor-induced CRF; two entities (polygons) are separated and the only dependency between states are within first-order.

- It mainly focuses on beneficial use of *outside* label as a medium delivering dependency between separated NEs.

- Focuses on label subsequence having $\langle entity, outside^+, entity \rangle$ pattern. (Figure 1 (a))

- Adds memory element to the hidden variables for the *outside* states (Figure 1(b)).

- The first *outside* label in an outside subsequence explicitly has a first-order dependency with its adjacent *entity*. If the first *outside* label tosses the information to the next, the information possibly flows forward.

- By induction process, the information of the first *entity* can flow through multiple *outside* labels to the second *entity* state (Figure 1(c)).

In the pre-induced CRF, the *outside* state with a memory element behaves as if an information transmission medium is delivering information about the presence or absence of the preceding entity forward. It is required to expand state set. States are collected and only entity states are selected. Multiplied *outside* state set is derived by multiplication of entity states and *outside* state. Expanded state set is consequently derived as a union of entity states and multiplied *outside* states.

Turning to the formulation, the conditional probability distribution of a label sequence $y$, given an observation $x$ in the CRF has a form as Eq.(1),

$$p(y|x) = \frac{1}{Z(x)} \cdot \prod_{t=1}^{T} exp\{\sum_{k=1}^{K} \theta_k f_k(y_t, y_{t-1}, x_t)\} \quad (1)$$

where $f_k$ is an arbitrary feature function having corresponding weight $\theta_k$, the $Z(x)$ is a partition function, and $t$ is time step (Sutton & McCallum, 2011). The feature function $f_k$ is generally indicator function that has value 1 only if the function is matched to a certain condition, otherwise 0. Transition factor in CRF has a form of function $f_{ij}(y, y', x) = \mathbf{1}_{\{y=i\}}\mathbf{1}_{\{y'=j\}}$, and observation factor has a form of a function $f_{io}(y, y', x) = \mathbf{1}_{\{y=i\}}\mathbf{1}_{\{x=o\}}$. Derived from Eq.(1), conditional probability distribution of the precursor-induced CRF takes a form as Eq.(2),

$$p(y|x,a) = \frac{1}{Z(x,a)} \cdot \prod_{t=1}^{T} exp\{\sum_{k=1}^{K} \theta_k f_k(y_t, y_{t-1}, x_t, a_t, a_{t-1})\} \quad (2)$$

where the variable $a$ is to store the induced state information, and the value of "$a_t$" is activated by the value of "$a_{t-1}$" and "$y_t$". Once the "$a_t$" is activated, the "$a_t$" eventually transmutes the value of "$y_t$."

This induction process eventually expands the original label value set. It produces newly induced *outside* states instead of the single *outside* state; for example, the process modifies an original label sequence $\langle A, O, \cdots O, B \rangle$ to $\langle A, A[O]^+, \cdots A[O]^+, B \rangle$. This transformation helps the CRF learn long-distance named entity transitions, even in the first-order form; from the modified example sequence, the model can learn label transition data $\{(A[O]^+, B)\}$ where entity $B$ depends on entity $A$ preceding itself. In terms of the number of newly produced states, when $N=|States|$ in the original first-order CRF (a state set consists of NE states and one outside state), this procedure introduces $N$ new states. (if the IOB2 tagging scheme (Tjong & Sang, 1995) is applied, $(N-1)/2 + 1$ new states are introduced).

To train the precursor-induced CRF, L-BFGS optimization method (Fei Sha & Fernando Pereira, 2003) and *l2*-regularization (Ng, 2004) are used as conventional first-order CRF exploits (Sutton & McCallum, 2011). Furthermore, the Viterbi algorithm is used for inference.

During training and inference, it is also required to treat the fragmented *outside* states as a single *outside* label in practice. First, a weight of an observation feature $f_{io}$ depends on the frequency of an observation as well as co-occurrence label data. Fragmenting a single *outside* state into multiple states may cause data-sparseness problems especially for observation features occurring within the fine-grained *outside* states in training time. To prevent the data sparseness problem derived by the precursor-induced CRF, observation factor $f_{io}(y,y',x)$ is customized as $(\mathbf{1}_{\{i \in \neg \text{Outside}, y=i\}} + \mathbf{1}_{\{i \in \text{Outside}\}}) \mathbf{1}_{\{x=o\}} \mathbf{1}_{\{y'=1\}}$. Second, the expected label alphabets in inference time are required to be matched to the label alphabets of given annotation. Therefore, the fragmented *outside* state reverts to the original *outside* label.

## 3 Experiments

All the experiments were performed by implementing both the original and precursor-induced CRF[1]. The activity refers to CRF implemented in MALLET (Andrew Kachites McCallum, 2002). To compare precursor-induced CRF with the original CRF in NER on the real-world clinical documents and biomedical literatures, three annotated NER corpus were used; i2b2 2012 NLP shared task data (Sun, Rumshisky, & Uzuner, 2013), discharge summaries of rheumatism patients at Seoul National University Hospital (SNUH), and JNLPBA 2004 Bio-Entity Recognition shared task data (Kim, Ohta, Tsuruoka, Tateisi, & Collier, 2004). The discharge summary of rheumatism patient corpus is built for this evaluation. This corpus consists of 200 electronic clinical documents where English and Korean words are jointly used for recording patient history. We used the division of training and test set provided by the i2b2 2012 and JNLPBA corpus in this evaluation. For the SNUH corpus, 10-fold cross validation was used.

Annotated named entities involved in the clinical NER evaluation are related to mentions describing the patient's history. In the i2b2 2012 corpus, *problem*, *test*, and *treatment* named entity classes are used. In the SNUH corpus, *symptom*, *test*, *diagnosis*, *medication*, and *procedure-operation* classes are used. The named entity classes in the biomedical NER evaluation are *DNA*, *RNA*, *protein*, *cell line*, and *cell type*.

In the i2b2 2012 training data, 9,942 entities have *outside* state precedence, and approximately 63.8% cases of them take a pattern $\langle entity, outside^+, entity \rangle$. Likewise, in SNUH corpus, 58.9% cases of NEs having *outside* precedence have a preceding named entity. Median value of the distance between consecutive entities tend to be within 3-4 in the datasets. The long distance dependency is restricted within a single instance (i.e., a sentence).

To perform NER evaluation, two types of feature families are used: (a) token itself and neighbor tokens in window size 3. In addition, morphologically normalized tokens are used together. (b) morphology features such as character prefix and suffix of length 2–4. Our *feature setting 1* uses the single feature family (a) and *feature setting 2* simultaneously uses both of the feature family (a) and (b). The reason for setting these simple feature configurations is for the purpose of reducing bias that the feature will affect the performance comparison of the models.

In order to compare the proposed model with the conventional CRF, both the first-order and the second-order CRF are used as baseline models.

---
[1] https://github.com/jinsamdol/precursor-induced_CRF

| Feature set | Model | i2b2 2012 | | | JNLPBA 2004 | | | SNUH | | |
|---|---|---|---|---|---|---|---|---|---|---|
| | | P | R | F | P | R | F | P | R | F |
| Set 1 (a) | first-order CRF | **77.04** | 63.88 | 69.84 | 66.27 | 62.61 | 64.39 | 82.42 | 73.81 | 77.85 |
| | second-order CRF | 74.72 | 63.35 | 68.56 | 68.03 | **63.53** | 65.70 | 83.27 | 75.45 | 79.14 |
| | pre-induced CRF | 76.25 | **65.13** | **70.25** | **69.23** | 62.54 | **65.71** | **83.73** | **75.83** | **79.57** |
| Set 2 (a)+(b) | first-order CRF | **75.73** | 67.09 | **71.15** | 67.38 | 69.43 | 68.39 | 84.83 | 80.30 | 82.49 |
| | Second-order CRF | 74.32 | 65.01 | 69.35 | 67.12 | 67.26 | 67.19 | 84.88 | 79.15 | 81.90 |
| | pre-induced CRF | 75.41 | **67.14** | 71.04 | **68.86** | **69.50** | **69.18** | **84.95** | **80.45** | **82.63** |

Table 1: Overall performance comparison. Shaded cells: baseline models. Bolded values: best performance within the comparison group (P: precision, R: recall, F: $F_1$-score).

The performance comparison result is shown in the Table 1. The result shows a tendency that precursor-induced (pre-induced) CRF leads to a slight performance improvement compared to both the first-order and second-order CRFs in most cases. However, the overall improvement is small.

Table 2 compares the elapsed time per iteration in parameter training for each model. The result shows that the second-order CRF takes quite more time than the first-order CRF to compute one training iteration. The pre-induced CRF takes 1.7 times more computation time than the first-order CRF in average. The pre-induced CRF takes significantly less time than the second-order CRF while the pre-induced CRF exploits longer label transition dependency than the second-order CRF.

These results indicate that the precursor-induced CRF, where long-distance dependency is introduced in CRF by label induction, slightly improves the effectiveness in clinical and biomedical NER while also significantly reducing computational cost rather than building second- or higher-order CRFs.

| Model | i2b2 | JNLPBA | SNUH |
|---|---|---|---|
| first-order | 3.97 | 39.39 | 5.44 |
| second-order | 30.34 | 497.15 | 87.49 |
| pre-induced | 6.55 | 69.78 | 9.08 |

Table 2: Elapsed training time (s/iteration)

## 4 Conclusion

The requirement utilizing high-order dependencies often holds in sequence labeling problems; however, second-order or higher-order models are considered computationally infeasible. Therefore, this study focuses on beneficial use of single *outside* label as a medium delivering long-distance dependency. The design of the precursor-induced CRF apparently allows precedent named entity information to pass through *outside* labels by induction, even when the model maintains a first-order template. Although the performance improvement is small in both the clinical and biomedical NER evaluations, this study has shown that the proposed design enables reduced computational cost in utilizing long-distance label dependency compared to the second-order CRF.

Evidence from this study suggests that the utilization of *outside* labels as precedent NE information transmission medium presumably can enhance the expressiveness of the CRF while keeping the first-order template. Considerable work is required to validate the model. For example, the validation of the precursor-induced CRF in deep neural architecture for NER, such as the LSTM-CRF neural architecture (Lample et al., 2016), will be worth performing in the future. In addition, validation of the model in various problems, such as NER in general domain (Tjong, Sang, & Meulder, 2003) and de-identification problem of personal health information in clinical natural language processing (Stubbs, Filannino, & Uzuner, 2017; Stubbs, Kotfila, & Uzuner, 2015), will be performed in the future study.

## Acknowledgements

This work was supported by the Basic Science Research Program through the National Research Foundation of Korea (NRF) funded by the Ministry of Education (No. NRF-2015R1D1A1A01058075), and also supported by a grant of the Korea Health Technology R&D Project through the Korea Health Industry Development Institute (KHIDI), funded by the Ministry of Health & Welfare, Republic of Korea (grant number : HI14C1277). Deidentified English clinical records used in this research were provided by the i2b2 National Center for Biomedical Computing funded by U54LM008748 and were originally prepared for the Shared Tasks for Challenges in NLP for Clinical Data organized by Dr. Ozlem Uzuner, i2b2 and SUNY.

# References

Andrew Kachites McCallum. (2002). MALLET: A Machine Learning for Language Toolkit. Retrieved March 27, 2013, from http://mallet.cs.umass.edu

Andrew McCallum, & Wei Li. (2003). Early Results for Named Entity Recognition with Conditional Random Fields , Feature Induction and Web-Enhanced Lexicons. In *Proceeding of CoNLL 2003* (pp. 188–191).

Cuong, N. V., Ye, N., Lee, W. S., & Chieu, H. L. (2014). Conditional Random Field with High-order Dependencies for Sequence Labeling and Segmentation. *ACM JMLR*, *15*, 981–1009.

Fei Sha, & Fernando Pereira. (2003). Shallow Parsing with Conditional Random Fields. In *Proceedings of the 2003 Conference of the North American Chapter of the Association for Computational Linguistics on Human Language Technology* (pp. 134–141).

Fersini, E., Messina, E., Felici, G., & Roth, D. (2014). Soft-constrained inference for Named Entity Recognition. *Information Processing and Management*, *50*(5), 807–819.

John Lafferty, Andrew McCallum, & Fernando Pereira. (2001). Conditional Random Fields : Probabilistic Models for Segmenting and Labeling Sequence Data. In *Proceedings of the 18th International Conference on Machine Learning 2001* (pp. 282–289).

Kim, J.-D., Ohta, T., Tsuruoka, Y., Tateisi, Y., & Collier, N. (2004). Introduction to the bio-entity recognition task at JNLPBA. *Proceedings of the International Joint Workshop on Natural Language Processing in Biomedicine and Its Applications - JNLPBA '04*, 70.

Lample, G., Ballesteros, M., Subramanian, S., Kawakami, K., & Dyer, C. (2016). Neural Architectures for Named Entity Recognition. In *Proceedings of NAACL-HLT 2016* (pp. 260–270).

Liu, Z., Tang, B., Wang, X., & Chen, Q. (2017). De-identification of clinical notes via recurrent neural network and conditional random field. *Journal of Biomedical Informatics*.

McDonald, R., & Pereira, F. (2005). Identifying gene and protein mentions in text using conditional random fields. *BMC Bioinformatics*, *6 Suppl 1*, S6. https://doi.org/10.1186/1471-2105-6-S1-S6

Ng, A. Y. (2004). Feature selection, L1 vs. L2 regularization, and rotational invariance. In *ICML 2004*.

Ratinov, L., & Roth, D. (2009). Design challenges and misconceptions in named entity recognition. In *Proceedings of the Thirteenth Conference on Computational Natural Language Learning* (pp. 147–155).

Sarawagi, S., & Cohen, W. W. (2005). Semi-Markov Conditional Random Fields for Information Extraction. In *Advances in neural information processing systems* (pp. 1185–1192).

Stubbs, A., Filannino, M., & Uzuner, Ö. (2017). De-identification of psychiatric intake records: Overview of 2016 CEGS N-GRID shared tasks Track 1. *Journal of Biomedical Informatics*, *75*, S4–S18.

Stubbs, A., Kotfila, C., & Uzuner, Ö. (2015). Automated systems for the de-identification of longitudinal clinical narratives: Overview of 2014 i2b2/UTHealth shared task Track 1. *Journal of Biomedical Informatics*, *58*, S11–S19.

Sun, W., Rumshisky, A., & Uzuner, O. (2013). Evaluating temporal relations in clinical text: 2012 i2b2 Challenge. *Journal of the American Medical Informatics Association : JAMIA*, 1–8.

Sutton, C., & McCallum, A. (2011). An Introduction to Conditional Random Fields. *Foundations and Trends in Machine Learning*, *4*(4), 267–373. https://doi.org/10.1561/2200000013

Tjong, E. F., & Sang, K. (1995). Representing Text Chunks, 173–179.

Tjong, E. F., Sang, K., & Meulder, F. De. (2003). Introduction to the CoNLL-2003 Shared Task : Language-Independent Named Entity Recognition. In *Proceedings of the seventh conference on Natural language learning at HLT-NAACL 2003* (pp. 142–147).

Ye, N., Lee, W. S., Chieu, H. L., & Wu, D. (2009). Conditional Random Fields with High-Order Features for Sequence Labeling. In *Advances in Neural Information Processing Systems* (pp. 2196–2204).